\documentclass[11pt,a4paper]{article}
\usepackage{coling2016}
\usepackage{times}
\usepackage{url}
\usepackage{latexsym}
\usepackage{wrapfig}

\usepackage{url}
\usepackage{booktabs}
\usepackage{graphicx}
\usepackage{color}
\usepackage{enumitem}

\title{Keystroke dynamics as signal for shallow syntactic parsing}

\author{Barbara Plank \\
  University of Groningen \\
  The Netherlands \\
  {\tt b.plank@rug.nl}}

\date{}

\begin{document}
\maketitle
\begin{abstract}
 Keystroke dynamics have been extensively used in psycholinguistic and writing research to gain insights into cognitive processing. 
But do keystroke logs contain actual \textit{signal} that can be used to \textit{learn} better natural language processing models?

We postulate that keystroke dynamics contain information about syntactic structure
that can inform shallow syntactic parsing. To test this hypothesis, 
we explore labels derived from keystroke logs as auxiliary task in a multi-task bidirectional Long Short-Term Memory (bi-LSTM). Our results show promising results on two shallow syntactic parsing tasks, chunking and CCG supertagging.
Our model is simple, has the advantage that data can come from distinct sources, and produces models that are significantly better than models trained on the text annotations alone. 
\end{abstract}

\section{Introduction}
\label{intro}

%
%
\blfootnote{
    %
    %
    \hspace{-0.65cm}  
    %
    %
     This work is licensed under a Creative Commons 
     Attribution 4.0 International Licence.
     Licence details:
     \url{http://creativecommons.org/licenses/by/4.0/}
    %
    %
}

As people produce text, they unconsciously produce loads of cognitive side benefit such as keystroke logs, brain activations or gaze patterns. 
However, natural language processing (NLP) hitherto almost exclusively relied on the written text itself. We argue that cognitive processing data contains potentially useful information beyond the linguistic signal and propose a novel source of information for shallow syntactic parsing, keystroke logs. 

\textit{Keystroke dynamics} concerns a user's typing pattern. When a person types, the latencies between successive keystrokes and their duration reflect the unique typing behavior of a person. Keystroke logs, the recordings of a user's typing dynamics, are studied mostly in cognitive writing and translation process research to gain insights into the cognitive load involved in the writing process. However, until now this source has not yet been explored to inform NLP models.

Very recent work has shown that cognitive processing data carries valuable signal for NLP. For instance, eye tracking data can inform sentence 
compression~\cite{klerke:ea:2016} and gaze is predictive for part-of-speech~\cite{barrett:ea:2015,barrett:ea:2016}.  

Keystroke logs have the distinct advantage over other cognitive modalities like eye tracking or brain scanning, that they are readily available and can 
be harvested easily, because they do not rely on any special equipment beyond a keyboard. Moreover, they are non-intrusive, inexpensive, and have the potential to offer continuous adaptation to specific users. Imagine integrating keystroke logging into (online) text processing tools.

We hypothesize that keystroke logs carry syntactic signal. Writing time between words can be seen as proxy of the planning process involved in writing, and thus represent structural information between words. To test our hypothesis, we evaluate a multi-task bidirectional Long-Short Term Memory (bi-LSTM) model that is---to the best of our knowledge---the first to exploit keystroke logs to improve NLP models. 
We test our model on two shallow syntactic tasks, chunking and CCG supertagging. The choice of tasks is motivated by the fact that writing research analyzes so-called \textit{bursts} of writing (i.e., consecutive spans of text, cf.\ Section~\ref{sec:background}), which are related to shallow syntactic annotation. 

To exploit the keystroke log information we model it as auxiliary task in a multi-task setup (cf.\ Section \ref{bilstm}). This setup has the advantage that the syntactic data 
and keystroke information can come from \textit{distinct} sources, thus we are not restricted to the requirement of jointly labeled data (a corpus with both annotations). 
Our exploratory evaluation shows that little keystroke data suffices to improve a syntactic chunker on out-of-domain data, and that keystrokes also aid CCG tagging.

\paragraph{Contributions} We are the first to use keystroke logs as signal to improve NLP models. In particular, the contributions of this paper are the following:
i) we present a novel bi-LSTM model that exploits keystroke logs as auxiliary task for syntactic sequence prediction tasks;
ii) we show that our model works well for two tasks, syntactic chunking and CCG supertagging, and 
iii) we make the code available at: \texttt{https://github.com/bplank/coling2016ks}.

\section{Keystroke dynamics}

We see keystroke dynamics as providing a complementary view on the data beyond the linguistic signal, which can be harvested easily and is particularly attractive to build robust models for out-of-domain setups. Keystroke logging data can be seen as an instance of \textit{fortuitous data}~\cite{Plank:2016:KONVENS}; it is side benefit of behavior that we want to exploit here. However, keystroke log is raw data, thus first needs to be \textit{refined} before it can be used. Our idea is to treat the duration of pauses before words as a simple sequence labeling problem.  

We first describe the process of obtaining auto-labeled data from raw keystroke logs, and then provide background and motivation for this choice. Section~\ref{bilstm} then describes our model, i.e., by solving the keystroke sequence labeling problem jointly with shallow syntactic parsing tasks (chunking and CCG supertagging) we want to aid shallow parsing.


\subsection{From keystroke logs to auxiliary labels}\label{sec:autolabel}
\label{sec:tolabels}

\begin{figure}\centering
\includegraphics[width=0.6\textwidth]{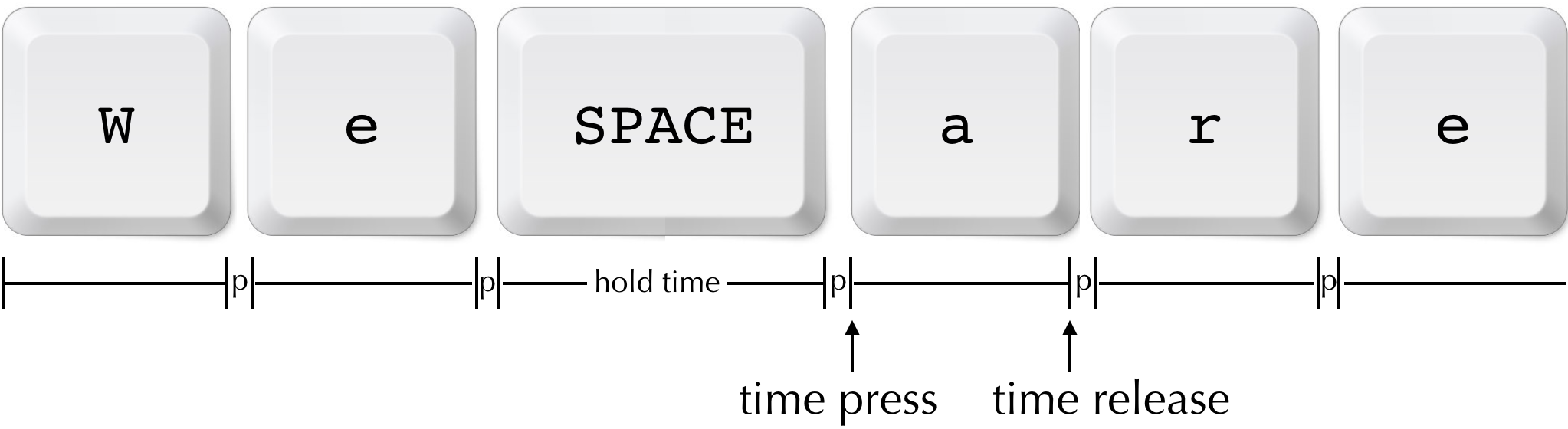}
\caption{Keystroke logging. $p$ are pauses between keystrokes.}
\label{fig:keyZ}
\end{figure}

While keystroke dynamics considers a number of timing metrics, such as \textit{holding time} and \textit{time press} and \textit{time release} between every keystroke ($p$ in Figure~\ref{fig:keyZ}), in this study we are only concerned with the pause preceding a word (i.e., the third $p$ in Figure~\ref{fig:keyZ}).\footnote{Figure inspired by the figure in~\cite{goodkind:ea:2015}.} We here use a simple tokenization scheme. Whitespace delimits tokens, punctuation delimits sentence boundaries. 

An example of pre-word pauses (in the remainder simply called \textit{pauses}) calculated from our actual keylog data is shown in Table~\ref{tbl:exampleLog}. If we take an arbitrary threshold of 500ms, the chunks indicated by the brackets are derived. This affirms that pre-word pauses carry constituency-like information. 

\begin{figure}\centering
\includegraphics[width=0.45\textwidth]{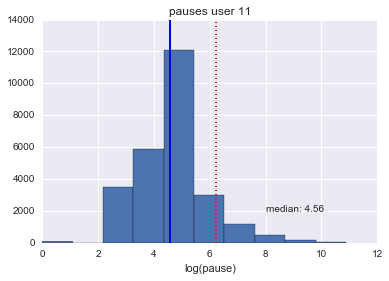}
\includegraphics[width=0.45\textwidth]{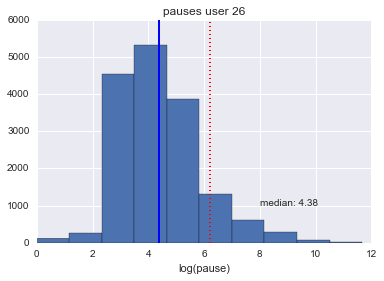}
\caption{Distribution of pauses for two users (plotted in log space). Red solid line: per-user median pause. Dotted line: arbitrary 500ms threshold. As can be seen from the plots, the users' typing dynamics differs.}
\label{fig:twousers}
\end{figure}

However, typing behavior of users differs, as illustrated in Figure~\ref{fig:twousers}. Hence, rather than finding a global metric we rely on per-user calculated aggregate statistics and discretize them to obtain auto-derived labels, as explained next.
\begin{table}
\resizebox{\textwidth}{!}{
\begin{tabular}{lccccccccccc}
\toprule
Token: & $[$ Coefficient & of & determination $]$ & $[$ is & a $]$  & $[$ measure & used & in $]$ & $[$ statisitcal &  model $]$ & $[$ analysis $]$ \\
\midrule
Pause (ms): & 0 & 96 & 496 & 30769 & 96 & 2144 & 96 & 80 & 2975 & 240 & 680 \\

\bottomrule
\end{tabular}
}
\caption{Example keystroke log for user 33 (including typo). If we segment the data using an arbitrary 500ms pre-word pause the chunks indicated by the brackets are obtained. To normalize over idiosyncrasies of users we use per-user average statistics  to obtain segments with auto-derived labels, see Section~\ref{sec:tolabels}.} 
\label{tbl:exampleLog}
\end{table}

We calculate $p$, the pause duration before a token, and bin it into the following categories, using BIO encoding, where $median$ is the per-user median and $mad$ the median absolute deviation. In this way, we automatically gather labels from keystrokes representing pause durations. 

In particular, we use the following discretization, i.e., a label for a token is calculated by:

\begin{tabular}{lll}
\noindent $label$ = & \texttt{<m} & if $p < median$;\\
& \texttt{<m+.5} & if $p < median+0.5 *mad$;\\
& \texttt{<m+1} & if $p < median+mad$;\\
& \texttt{>m1} & else;\\
&\texttt{O} & for punctuation symbols.\\
\end{tabular}

The label is further enriched with a prefix in BIO encoding style, motivated by the fact that we want to model spans of information. Punctuation symbols are treated as \texttt{O}, because due to their location at boundary positions the pause information varies highly. We leave treating punctuation separately as future work. 
~\newcite{klerke:ea:2016} use a related encoding scheme to discretize fixation durations obtained from eye tracking data, however, in contrast to them we here use median-based measures which are better suited for such highly skewed data~\cite{leys2013detecting}. An actual example of automatically labeled keystroke data is given in Table~\ref{tbl:example}. 

\begin{table}[ht!]\centering
\begin{tabular}{c|c|c|c|c|c|c}
\toprule
\texttt{B-<m} & \texttt{B-<m+1} &  \texttt{B-<m} & \texttt{I-<m} & \texttt{B-<m+.5} & \texttt{I-<m+.5} & \texttt{B->m+1}\\

\midrule
the & closer & the & number & is & to & 1\\
\bottomrule
\end{tabular}
\caption{Example auto-derived keystroke annotation.}
\label{tbl:example}
\end{table}


\subsection{Background} 
\label{sec:background}
The major scientific interest in keystroke dynamics is that it provides a non-intrusive method for studying cognitive processes 
involved in writing. Keystroke logging has developed to a promising tool in writing research~\cite{sullivan:lindgren:2006,nottbusch:ea:2007,wengelin:2006,vanwaes:ea:2009,baaijen:ea:2012}, where
time measurements---pauses, bursts and revisions (described below)---are studied as traces of the recursive nature of the writing process.


In its raw form, 
keystroke logs contain information on which key was pressed for how long (key, time press, time release). This data is then used to calculate between keystroke pause durations, such as pre-word pauses. 
It has been shown that pauses reflect the planning of the unit of text itself~\cite{baaijen:ea:2012} and that they correlate with clause and sentence boundaries~\cite{spelman:sullivan:2006}.
Writing research is interested in \textit{bursts} of writing, defined as consecutive chunks of text produced and defined by a 2000ms time of inactivity~\cite{wengelin:2006}, or revisions. 
Such a cutoff is rather arbitrary~\cite{baaijen:ea:2012}, and from our own experience results in long chunks.  Taking writing research as a starting point,
we postulate that keystrokes contain further fine-grained information that help identify syntactic chunks. We aim at a finer-grained representation, and transform user-based average statistics 
into automatically derived labels (cf.\ above). 

We notice that the literature defines different ways to define a pause. 
~\newcite{goodkind:ea:2015}, coming from a stylometry background, use the difference between release time of the previous key and the timepress of the current key to calculate pre-word pause duration.\footnote{Goodkind sets negative pause durations (which can arise in this setup) to 0 (personal communication).}
In contrast, writing research~\cite{wengelin:2006,vanwaes:ea:2009,baaijen:ea:2012} defines pauses as the start time of a keystroke until the start time of the next keystroke. 
We experimented with both types of pause definitions, and found the former slightly more robust, hence we use that to calculate pauses throughout this paper.


\begin{figure}\centering
\includegraphics[width=0.35\textwidth]{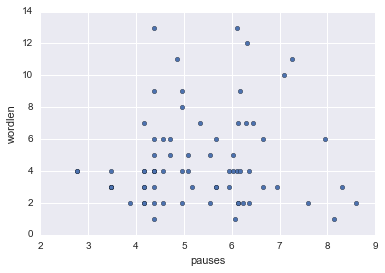}
\includegraphics[width=0.35\textwidth]{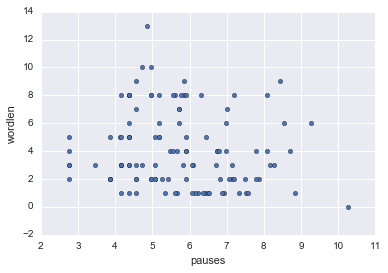}
\caption{left: Word pauses length vs word length (left: user 7, right: user 3; Pearson $\rho=0.08$, and $\rho=-0.12$)} 
\label{fig:plotsWordLen}
\end{figure} 

\begin{figure}\centering
\includegraphics[width=0.69\textwidth]{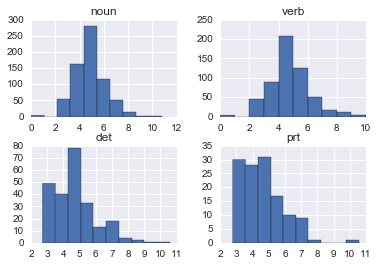}
\caption{Word pause length distribution per part-of-speech (user 5).} 
\label{fig:plotsPOS}
\end{figure}

In order to get a better feel of word pause durations, we examine various properties of them. First, do we need to normalize pauses for word length? ~\newcite{goodkind:ea:2015} found a linear relationship between 
pre-word pauses and word length in their dataset. We calculated the correlation between word length and pauses in our dataset, but could not observe such a relation in our data 
(cf.\ Figure~\ref{fig:plotsWordLen}; plots for the other participants looks similar). Even if we break the data down by POS and calculate per-POS wise correlations we found no relation between pause duration and word length.\footnote{POS annotations were obtained by looking up the 
possible tag of a token in English wiktionary~\cite{Li:ea:12}.
} Hence we do not normalize word pause durations. In addition, Figure~\ref{fig:plotsPOS} plots pauses for various part-of-speech, showing that function POS (determiner, particles) are preceded by shorter pauses than content POS (we obtain similar plots for other participants).

Second, keystroke logs are presumably idiosyncratic, can we still use it? In fact, user keystroke biometrics are successfully used for author stylometry 
and verification in computer security research~\cite{stewart:ea:2011,monaco2013behavioral,locklear:ea:2014}. However, also eye tracking data like scanpaths (the resulting series of fixations and saccades in eye tracking) are known to be idiosyncratic~\cite{kanan:ea:2015}. Nevertheless it has been shown that gaze patterns help to inform NLP~\cite{barrett:ea:2015,klerke:ea:2016}. We believe this is also the case for biometric keystroke logging data.




\section{Tagging with bi-LSTMs}\label{bilstm}

We draw on the recent success of bi-directional recurrent neural network (bi-RNNs)~\cite{graves:schmidhuber:2005}, in particular Long Short-Term Memory (LSTM) models~\cite{Hochreiter:Schmidhuber:97}. They read the input sequences twice, in both directions. Bi-LSTM have recently successfully been used for a variety of tasks~\cite{Collobert:ea:2011natural,ling:ea:2015,wang:ea:2015:arxiv,Huang:ea:15,dyer:ea:2015,ballesteros:ea:2015,kiperwasser:goldberg:2016,Liu:ea:15}. For further details, see \newcite{goldberg-primer} and \newcite{cho-primer}.  

\begin{figure}\centering 
\includegraphics[width=0.95\textwidth]{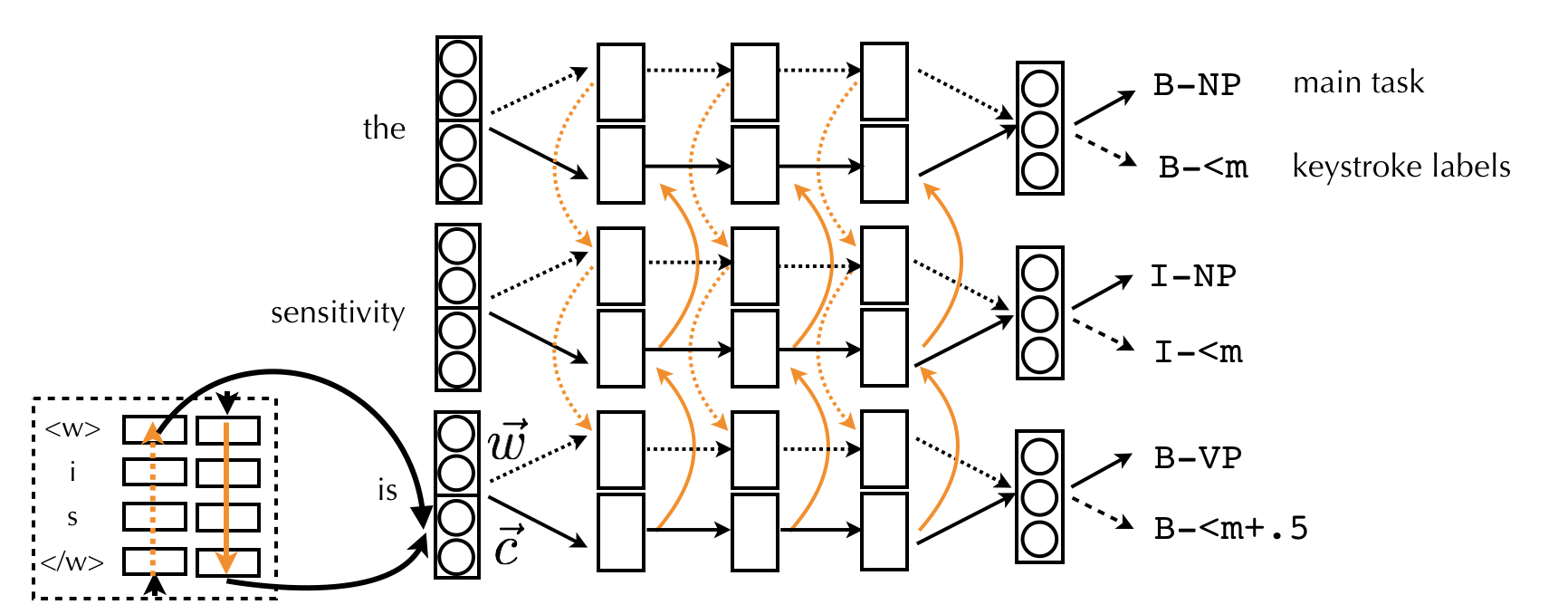}
\caption{Hierarchical Bi-LSTM with 3 stacked layers using word $\vec{w}$ and characters $\vec{c}$ embeddings.}
\label{fig:model}
\end{figure}

\subsection{Bidirectional Long-Short Term Memory Models}

Our model is a a hierarchical bi-LSTM as illustrated in Figure~\ref{fig:model}. It takes as input word embeddings  $\vec{w}$  concatenated with character embeddings obtained from the last two states (forward, backward) of running a lower-level bi-LSTM on the characters. Adding character representations as additional information has been shown to be effective for a number of tasks, including parsing and tagging~\cite{ballesteros:ea:2015,gillick:ea:2015,plank:ea:2016}.

In more detail, our model is a context bi-LSTM taking as input word embeddings $\vec{w}$. Character embeddings $\vec{c}$ are incorporated via a 
hierarchical bi-LSTM using a sequence bi-LSTM at the lower level~\cite{ballesteros:ea:2015,plank:ea:2016}. The character representation is concatenated with the (learned) word
embeddings $\vec{w}$ to form the input to the context bi-LSTM at the upper layers. 

For the hidden layers, we use stacked LSTMs with $h$=3 layers. The 3-layer bi-LSTM and lower-level character bi-LSTM represents the shared structure between tasks. From the topmost ($h$=$3$) layer labels for the different tasks (e.g., chunking, pauses) are predicted using a softmax. In Figure~\ref{fig:model}, the main task (chunking or CCG tagging) is represented by the solid arrow, the auxiliary task (keystroke logs) is indicated by the dashed arrow.

During training, we randomly sample a task and instance, and backpropagate the loss of the current instance through the shared deep network. In this way, we learn a joint model from distinct sources. Note that we also experimented with predicting the pause durations 
at lower levels ($h$=1), motivated by having lower-level tasks at lower layers in the network~\cite{soegaard:goldberg:2016}, however, we found the setup with both tasks at the outer layer more robust. Predicting all tasks at the outermost layer is the most commonly used form of multi-task learning in neural networks~\cite{caruana1998multitask,Collobert:ea:2011natural}.

\section{Experiments} 

We implement our model in \texttt{CNN/pycnn}.\footnote{\url{https://github.com/clab/cnn}} For all experiments, we use the same hyperparameters, 
set on a held-out portion of the CoNLL 2000 data, i.e., SGD with cross-entropy loss, no mini-batches, 30 epochs, default learning rate (0.1), 64 dimensions for 
word embeddings, 100 for character embeddings, random initialization for all embeddings, 100 hidden states, $h=3$ stacked layers, Gaussian noise with $\sigma$=0.2. As training is stochastic, we use a fixed seed
throughout (chosen and fixed upfront). 
No further unlabeled data is considered.


\paragraph{Datasets}

An overview of the syntactic datasets considered in this paper is given in Table~\ref{tbl:overviewstats}. For chunking, we use the original CoNLL data~\cite{conll2000} from WSJ (WSJ sections 15-18 as training data and section 20 as test data, containing 8936 and 2012 sentences, respectively).\footnote{\url{http://www.cnts.ua.ac.be/conll2000/chunking/}} 
For testing we take out-of-domain data whenever available, to test the adaptability of the method to noisy out-of-domain   data. For chunking we use Twitter data from~\newcite{ritter:2011} (all, 2364 tweets) and~\newcite{Foster:ea:11} (250 sentences), converted to chunks \cite{Plank:ea:14}.

The CCG supertagging data
also comes from WSJ (39604 training and 2407 test sentences). We unfortunately do not have access to out-of-domain test data, hence use the CCG tagging test set. 

\begin{table}[ht!]\centering
\begin{tabular}{lrrr}
\toprule
 sentences &\textsc{Train} & \textsc{Dev} & \textsc{Test}\\
 \midrule
\textsc{CoNLL 2000} & 8936 & -- &2012 \\   
\textsc{Foster} & -- & 269 & 250\\
\textsc{Ritter} & -- & -- & 2364\\
\midrule
\textsc{CCG} & 39604 & 1913 & 2407\\
\bottomrule
\end{tabular}
\caption{Statistics on the data sets}
\label{tbl:overviewstats}
\end{table}

The keystroke logging data stems from students taking an actual test on spreadsheet modeling in a university course~\cite{stewart:ea:2011,monaco2013behavioral}. 
The advantage of this dataset is that it contains free-text input.\footnote{In contrast to \url{http://www.casmacat.eu/} data that logs revisions from MT post-editing.}  
We used data from 38 users,\footnote{Disregarding users due to issues with logging~\cite{stewart:ea:2011}.} which produced on average 250 sentences.
The data totals to 7699 sentences.

To evaluate our models we use standard evaluation measures computed with \texttt{conlleval.pl} with default parameters, i.e., we report F1 on chunks and accuracy on CCG tags. Statistical significance is computed using the approximate randomization test~\cite{noreen} using $i=1000$ iterations and $p$-values are reported~\cite{sogaard2014s}.

%
%

\subsection{Results}

\begin{table}\centering

\begin{tabular}{lc|lc}
\toprule
Chunking     & F1 & CCG tagging & Accuracy\\
     \midrule
Our model & 93.21 & Our model & 92.41\\ 
\newcite{suzuki:isozaki:2008} & 93.88 & \newcite{xu:ea:2015}  & 93.00\\
\bottomrule
\end{tabular}
\caption{Baseline model, comparison to existing systems}
\label{tbl:baselines}
\end{table}

\paragraph{Baseline model} Both or baseline models are comparable to prior work, while being simpler. The results are summarized in Table~\ref{tbl:baselines}.
Our chunking baseline achieves an $F1$ of 93.21 on CoNLL, compared to the $F1$ of 93.88 of~\newcite{suzuki:isozaki:2008}, who use a CRF and gold POS tags. We do not use any POS information. A similar bi-LSTM achieves 93.64~\cite{Huang:ea:15}, however, additionally uses POS embeddings. 
Our baseline CCG supertagging model achieves 92.41, compared to the more complex model by~\newcite{xu:ea:2015} achieving an accuracy of 93.00. Very recently even higher accuracies were reported, e.g.~\cite{vaswani-EtAl:2016:N16-1}, however, in this exploratory paper we are interested in examining whether we find signal in keystroke data, and are not interested in beating the latest state-of-the-art.


\begin{table}[ht!]\centering

\begin{tabular}{llll|l}
\toprule
           & \textsc{Foster.dev} &  \textsc{Foster.test}  &  \textsc{Ritter} & \textsc{CCG}\\
           \midrule
Baseline & 73.93  & 73.61  &  66.65 & 92.41 \\
+\textsc{Pause} & \textbf{74.63}$^\dagger$ & \textbf{74.32}$^\dagger$ & \textbf{66.91}$^\dagger$ & \textbf{92.62}$^\dagger$ \\
\midrule
$p$-values & $<$0.01 & $<$0.01 & $<$0.01 &  $<$0.048\\
 \bottomrule
\end{tabular}
\caption{Chunking results (F1, +Pause is average over 38 participants) 
and CCG accuracy (using all pause data at once). Results marked with $^\dagger$ are  significantly better than the corresponding baseline using a randomization test with $i=1000$ iterations; $p$-values provided in row below.}
\label{tbl:results}
\end{table}
\paragraph{Keystroke pauses} The aim of our experiments is to gauge whether through joint learning of shallow syntax and pause duration the system learns to generalize over the pause information and thus aids the syntactic signal. 

The results in Table~\ref{tbl:results} support our hypothesis that keystroke dynamics contains useful information for chunking. We here report the average over models trained on a per-user basis, i.e., 38 participants.  
The results show that overall F1 chunking score improves over all datasets. For instance on the Ritter data, for 25/38 participants using their keystroke information as auxiliary task helps to 
improve overall chunking performance. However, if we combine all data and train a single model, performance degrades on chunking. We attribute this effect to the fact that the chunking data is relatively small, and higher 
amounts of keystroke data show signs of overfitting.  In fact, similar effects have been shown in a multi-task machine translation and parsing setup~\cite{luong:ea:2016}, where mixing coefficients were used to downplay the 
importance of the auxiliary parsing data that otherwise swamped the main task data.  We leave 
examining task-specific weights for the loss for future work.

In contrast in CCG tagging, where we have more training data, we see a positive effect of using keystroke data when training a model that uses all keystroke data at once (concatenation of all keystroke data from all users), see last column in Table~\ref{tbl:results}. Note that all results in Table~\ref{tbl:results} are significant.

\begin{table}[ht!]\centering
\begin{tabular}{lllllll}
\toprule
            & & \textsc{Foster.dev} &  \textsc{Foster.test}  &  \textsc{Ritter} \\
            \midrule
Baseline             & NP & 72.18 &  71.41 & 61.76\\
              & VP & 70.25 &  73.44 & 75.13\\
              & PP & 93.25 & 91.85 & 89.05 \\
             \midrule
             
 +\textsc{Pause}             & NP &  \textbf{73.99} & \textbf{72.77} & \textbf{62.60}\\
              & VP & 69.88 & \textbf{74.93} & 75.05\\
              & PP &  93.24 & 90.82 & 88.87 \\
              \bottomrule
\end{tabular}
\caption{Chunking results per label. }
\label{tbl:resultsPerLab}
\end{table}

\begin{table}[ht!]\centering
\begin{tabular}{lll}
\toprule
\textsc{Label} & \textsc{Baseline} &  +\textsc{Pause}  \\ 
\midrule
\texttt{N} &  97.20 & \textbf{97.27}\\ 
\texttt{N/N} &  96.38 &  \textbf{96.62} \\
\texttt{NP[nb]/N} & 99.02 & \textbf{99.03} \\
\texttt{(NP\textbackslash{}NP)/NP} & 88.41 &  \textbf{88.95} \\
\texttt{NP}  & 97.06 & \textbf{97.63}  \\
\texttt{PP/NP} & 72.60 &  \textbf{73.64} \\
\texttt{((S\textbackslash{}NP)\textbackslash{}(S\textbackslash{}NP))/NP} &  72.83 &  \textbf{74.07} \\
\texttt{conj} & 98.62 & \textbf{98.67}\\
              \bottomrule
\end{tabular}
\caption{CCG tagging results per label (most frequent non-punctuation labels shown). }
\label{tbl:resultsPerLabCCG}
\end{table}

\section{Discussion} 

To gain better insights into what the model has learned, 
Table~\ref{tbl:resultsPerLab} provides the per-label breakdown for chunking,  Table~\ref{tbl:resultsPerLabCCG}  for CCG tagging. Most of the improvements come from noun phrases (NP) chunks. From manual inspection we determine that the model improves particularly on non-conventional spelling and fragmented noun phrases typical for Twitter, see examples given in Table~\ref{tbl:resultEx}.

As Table~\ref{tbl:resultsPerLab} shows, keystroke data also helps for verb phrases on one dataset. The current encoding is not so beneficial for PPs. Pauses before prepositions are short, as illustrated in Figure~\ref{fig:plotsPOS}, and pauses often fall within segments in the auxiliary annotation, while prepositions constitute separate tokens in chunking. Hence, it is unsurprising that the model fares worse on PPs. 

We believe that our pause encoding mainly captures structural information between words, less morphosyntactic information itself, i.e., that pauses are more informative of syntactic structure than of part-of-speech. This intuition is in fact confirmed by initial experiments on POS tagging (UD English), which are less promising. We observe small improvements for low amounts of auxiliary data, however, they are not significant. Thus keystrokes seem to capture mostly structural shallow syntactic information, as confirmed in our experimental evaluation. However, this is only a first exploration, with one way of using keystroke logging data, but given our promising results, further experiments are warranted.

\begin{table}[ht!]\centering
\begin{tabular}{lllllll}
\toprule
\textsc{Token}            & \textsc{Gold}  & \textsc{Baseline} & \textsc{Model} \\
            \midrule
Auburn  &  B-NP &   I-NP & B-NP\\
party & I-NP  &  I-NP & I-NP\\
at & B-PP   & B-PP & B-PP\\
\midrule
Spurs & B-NP & B-NP & B-NP\\
v      & B-NP & I-NP & B-VP\\
Man  & B-NP  &  I-NP & B-NP\\
utd & I-NP  &  B-VP & I-NP\\
\midrule
sounds &   B-VP   &  B-VP   &     B-VP \\
bithcy  & B-ADJP & B-VP &                                         B-ADJP\\
 \bottomrule
\end{tabular}
\caption{Selected examples from the Twitter datasets.}
\label{tbl:resultEx}
\end{table}

\section{Related Work}

Keystroke logging has developed into a promising tool for research into writing~\cite{wengelin:2006,vanwaes:ea:2009,baaijen:ea:2012}, as time measurements can give insights into cognitive processes involved in writing~\cite{nottbusch:ea:2007} or translation studies. In fact, most prior work that uses keystroke logs focuses on experimental research. For example,~\newcite{hanoulle2015translation} study whether a bilingual glossary reduces the working time of professional translators. They consider pause durations before terms extracted from keystroke logs and find that a bilingual glossary in the translation process of documentaries reduces the translators' workload. Other translation research has combined eye-tracking data with keystroke logs to study the translation process~\cite{carl2016measuring}. An analysis of users' typing behavior was studied by \newcite{baba2012spelling}. They collect keystroke logs of online users describing images to measure spelling difficulty. They analyzed corrected and uncorrected spelling mistakes in Japanese and English and found that spelling errors related to phonetic problems remain mostly unnoticed.

\newcite{goodkind:ea:2015} is the only study prior to us that use keystroke loggings in NLP. In particular, they investigate the relationship between pre-word pauses and 
multi-word expressions and found within MWE pauses vary depending on cognitive task. We take a novel approach and \textit{learn} keystroke patterns and use them to inform shallow syntactic parsing. 

A recent related line of work explores eye tracking data to inform sentence compression~\cite{klerke:ea:2016} and induce part-of-speech~\cite{barrett:ea:2015}. Similarly, there are recent studies that predict fMRI activation from reading~\cite{wehbe:ea:2014} or use fMRI data for POS induction~\cite{bingel:ea:2016}.
The distinct advantage of keystroke dynamics is that it is easy to get, non-expensive and non-intrusive.

\section{Conclusions}

Keystroke dynamics contain useful information for shallow syntactic parsing. Our model, a bi-LSTM, integrates keystroke data as auxiliary task, and outperforms models
trained on the linguistic signal alone. We obtain promising results for two  syntactic tasks, chunking and CCG supertagging. This warrants many directions for future research, e.g., 
using information from the non-linear writing process, which we here disregarded (e.g., revisions), evaluating on other languages and going to the full parsing task.

\section*{Acknowledgements}
I would like to thank Veerle Baaijen for insightful discussions, and the three anonymous reviewers as well as H\'{e}ctor Mart\'{i}nez Alonso, Maria Barrett and Zeljko Agi\'{c} for comments on earlier drafts of this paper. I acknowledge the Center for Information Technology of the University of Groningen for their support and for providing access to the Peregrine high performance computing (HPC) cluster, as well as NVIDIA corporation for supporting my research.
\bibliographystyle{acl}
\bibliography{biblio}

\end{document}